\documentclass[letterpaper, 10 pt, conference]{ieeeconf}   

\IEEEoverridecommandlockouts                              

\usepackage[table,xcdraw]{xcolor}
\usepackage{todonotes}
\usepackage{booktabs}
\usepackage{cite}
\usepackage{amsfonts}

\usepackage{float}
\let\labelindent\relax
\usepackage{enumitem}
\usepackage{epsfig}
\usepackage{epstopdf}
\usepackage{amsmath,hyperref}
\usepackage[linesnumbered, ruled, vlined]{algorithm2e}

\makeatletter
\newcommand*{\rom}[1]{\expandafter\@slowromancap\romannumeral #1@}
\makeatother

\usepackage{xcolor}

\SetCommentSty{mycommfont}

\title{\LARGE \bf
Energy-Efficient Motion Planner for Legged Robots}

\author{Alexander Schperberg$^{{\dagger}{*}}$, Marcel Menner$^{\ddagger}$, and Stefano Di Cairano$^{\dagger}$%
    \thanks{$^*$Corresponding author}
    \thanks{$^{\dagger}$Alexander Schperberg, and Stefano Di Cairano are with Mitsubishi Electric Research Laboratories (MERL), Cambridge, MA, 02139, USA. {\tt\small \{schperberg, dicairano\}@merl.com}}
    \thanks{$^{\ddagger}$Marcel Menner was with Mitsubishi Electric Research Laboratories (MERL) during part of the work of this paper, Cambridge, MA, 02139, USA. {\tt\small \{menner\}@ieee.org}}
}

\begin{document}

\maketitle
\thispagestyle{empty}
\pagestyle{empty}

\begin{abstract}
We propose an online motion planner for legged robot locomotion with the primary objective of achieving energy efficiency. The conceptual idea is to leverage a placement set of footstep positions based on the robot's body position to determine when and how to execute steps. In particular, the proposed planner uses virtual placement sets beneath the hip joints of the legs and executes a step when the foot is outside of such placement set. Furthermore, we propose a parameter design framework that considers both energy-efficiency and robustness measures to optimize the gait by changing the shape of the placement set along with other parameters, such as step height and swing time, as a function of walking speed. We show that the planner produces trajectories that have a low Cost of Transport (CoT) and high robustness measure, and evaluate our approach against model-free Reinforcement Learning (RL) and motion imitation using biological dog motion priors as the reference. Overall, within low to medium velocity range, we show a 50.4\% improvement in CoT and improved robustness over model-free RL, our best performing baseline. Finally, we show ability to handle slippery surfaces, gait transitions, and disturbances in simulation and hardware with the Unitree A1 robot. 
\end{abstract}


\section{Introduction}
Legged robots excel on uneven terrain and in human-centric environments, but their efficiency is hindered by energy losses from foot-ground impacts, which accelerate battery depletion and increase motor heat, reducing operation time. Thus, formulating new energy conservation strategies are worthwhile endeavors. Hardware solutions such as Ranger~\cite{ranger_robot} and ANYmal~\cite{anymal_robot}, reduce energy consumption using passive dynamics and mechanical springs. Software-based solutions, the focus of this work, can include various motion planning strategies, including optimization~\cite{schperberg2022auto}, Reinforcement Learning (RL)~\cite{RL_energy}, and vision-based methods~\cite{vision_energy}. Although energy-efficient rewards or cost terms can be incorporated into the optimization or learning process, it is often challenging to assign significant weight to these terms without encountering local optima issues, where the robot fails to follow the desired velocity commands or avoids movements entirely. 

Our approach achieves energy-efficient locomotion using a simple, geometry-based footstep planner that constrains motion based on fundamental physical principles. Elliptical placement sets beneath each hip dictate when and where steps occur. The planner’s parameters—ellipse shape, swing time, and step height—are optimized to minimize the Cost of Transport (CoT) and maximize robustness to avoid singularity and favor non-slipping configurations. These parameters can be derived offline or adjusted through RL for improved disturbance handling. The proposed planner produces a more natural gait, where the robot moves its feet only when necessary, as defined by the placement set.

\begin{figure}[t]
    \centering
    \includegraphics[width=0.99\columnwidth]{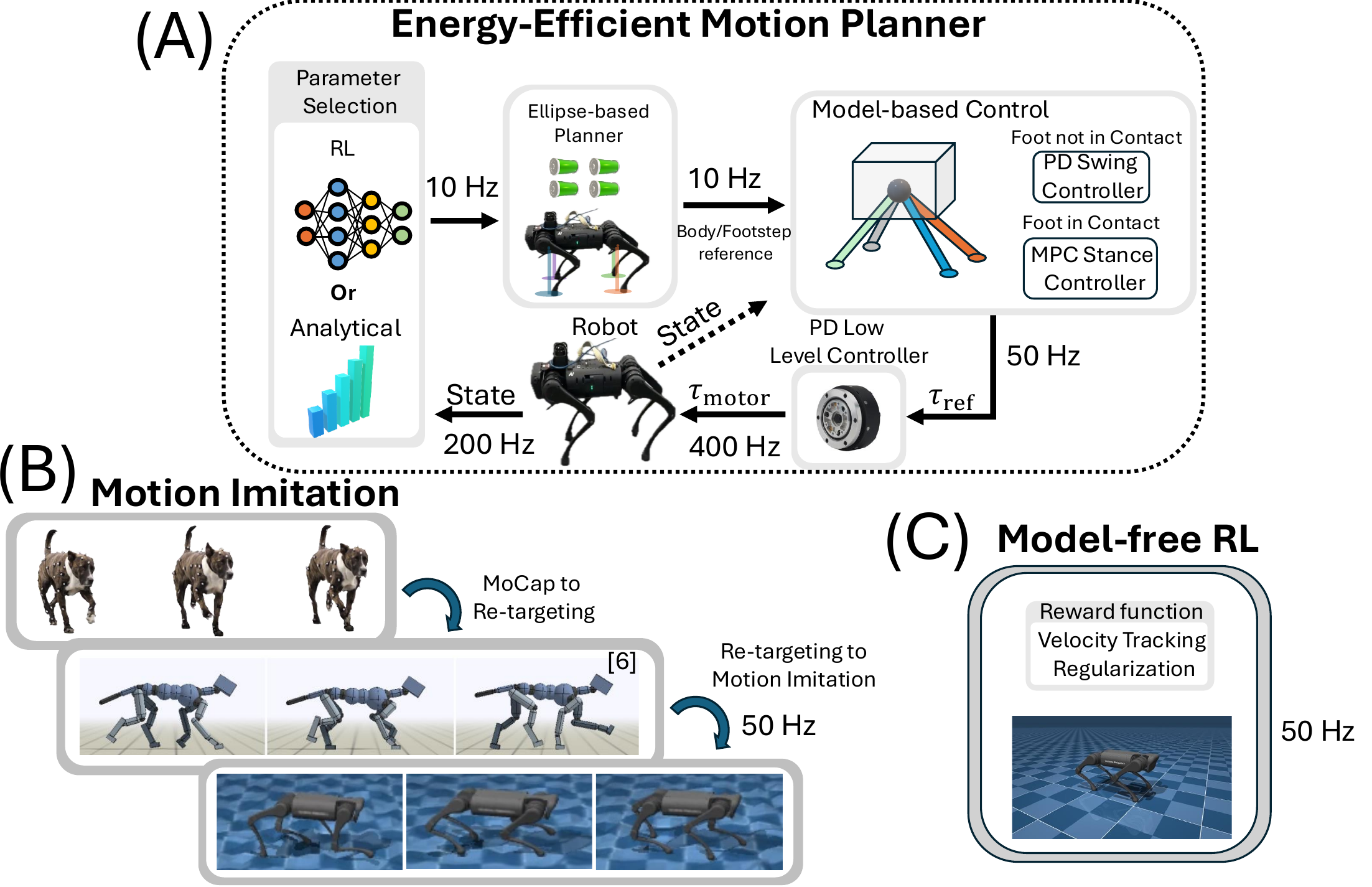} 
    \caption{We compare and evaluate our energy-efficient motion planner (A) (Sec. \ref{footstep_planner}), with motion imitation (B) (motion dataset from \cite{RoboImitationPeng20}) and model-free reinforcement learning (C) baselines.
    }
    \label{fig:robot}
\end{figure}

Although natural gaits can also be achieved by imitating animal motion \cite{RoboImitationPeng20}, to the best of our knowledge, there has not been a clear analysis to evaluate whether imitating motion derived from biological animals is actually more energy-efficient for lower degree of freedom robots. Thus, in this work, we use CoT and robustness measures to compare our method against not only typical model-free RL approaches \cite{Lee_2020}, but also against motion imitation, where the reference motions are derived from real animal movements (see Fig. \ref{fig:robot}). 

Overall, we present the following \textbf{contributions}:
\begin{enumerate}
    \item An online motion planning framework that naturally determines the gait by minimizing energy consumption using ellipses beneath the hip joints. Simulation and hardware experiments are presented.
    \item The ellipses, along with step height and swing time of each leg are designed based on a parameter study that considers CoT and robustness measures. This study is achieved analytically and also through exploration of parameters using RL. Both are compared against each other for efficacy.
    \item Our approach is compared against baselines, including model-free RL, and motion imitation using real biological dog motion priors as the reference.
\end{enumerate}

\begin{figure*}[!t]
    \centering
    \includegraphics[width=6.8in]{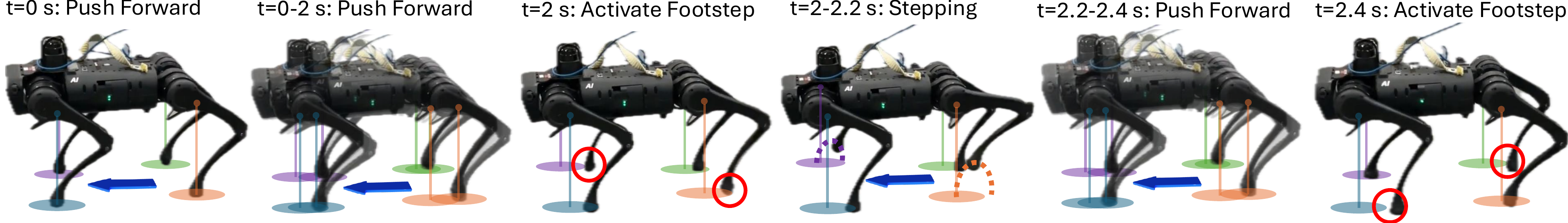}
    \caption{We show the process of our energy-efficient motion planner. With a desired forward velocity, from $t=0-2$ s, the robot's body pushes forward using the MPC stance controller from Sec. \ref{sssec:stance}, until one or more legs are outside of the ellipse (red circle), which activates the footstep as shown at $t=2$ s. The robot then takes a step, $t=2-2.2$ s, and the process repeats. Note, the center position of the ellipse is beneath the hip joint at all times.}  
    \label{3d_sequence}
\end{figure*}

\subsection{Related Works}
Motion planning is a fundamental requirement for legged robot locomotion. Several approaches for stable motion have been proposed such as model-based (Sec. \ref{related_works_model_based}) and learning-based methods including RL and motion imitation (Sec.\ref{related_works_learning}). Efficacy of these methods are typically evaluated primarily for velocity tracking, and only a few for energy-efficient locomotion (Sec. \ref{related_works_energy}).
\subsubsection{Model-based}
\label{related_works_model_based}
The Raibert controller~\cite{Raibert} provides stable trajectories for legged robots that is computationally efficient, but not suitable for rough terrain nor considers energy-efficiency. For rough terrain, optimization-based techniques can be used, optimizing for the Zero-Moment Point (ZMP) through the Linear Inverted Pendulum approach~\cite{kajita_ZMP}. Trajectory Optimization (TO) is employed if footsteps are not specified \textit{a priori}, which helps expand the robot's feasible workspace, and optimizes over contact forces using more complex centroidal dynamics~\cite{trajOpt}. However, TO methods are computationally expensive and difficult to apply for online-replanning. 

\subsubsection{Reinforcement Learning and Motion Imitation}
\label{related_works_learning}
To optimize over a high-precision model while being computationally efficient, model-free RL methods have become popular~\cite{RL_2,Lee_2020}, such as using a student/teacher policy in~\cite{Lee_2020}. However, these methods require extensive hyperparameter tuning, heavy reward engineering, and safety consideration is less relevant as the underlying network output is difficult to understand. Another approach is through motion imitation, such as Imitation Learning (IL)~\cite{imitation_deepMimic,RoboImitationPeng20}. For example, replicating motion patterns observed in biological locomotion data collected from a real dog was proposed in~\cite{RoboImitationPeng20}. By following such motion priors, the need to enforce walking behaviors such as achieving a certain step height, forward progress, or motion smoothness and so on, is greatly reduced, simplifying the number of reward terms necessary for legged locomotion. Although prior works using RL or IL can, and often do, include energy-efficient terms, the resulting policies tend to perform poorly if weights on these terms are too large. Also, these methods evaluate success based primarily on overall velocity tracking performance, not on energy-efficiency, as we do in this work.  

\subsubsection{Optimizing for Energy-Efficiency}
\label{related_works_energy}
Only few works primarily focus on energy-efficient locomotion to help reduce power consumption of the robot's actuators. One such example can be found in~\cite{energy_1}. Unlike our work, in~\cite{energy_1}, the step-height needs to be pre-determined, and the gait timings are a function of a pre-defined distance instead of being modifiable during operation. Although our work does not employ vision data as shown in~\cite{energy_2}, by considering energy-efficiency for various step heights implicitly provides information on which postures are optimal (energy wise) on uneven terrain. Similar to~\cite{energy_2}, we also consider slipping using force ellipsoids. 

Other works aim at energy-efficiency by either optimizing locomotion parameters~\cite{energy_4}, employing policy gradient RL~\cite{RL_energy}, or with adaptive control methods~\cite{Koco2014LocomotionCF}. However, these works do not consider the effects of minimizing energy-consumption while accounting for gait transitions which is computationally costly if using a programming solver~\cite{model_based_energy1,model_based_energy2}. In contrast, here we address the decision at which forward velocity it becomes more energy-efficient to transition from a gait to another, e.g., walking to trot. The proposed method is also general enough not to need the exact order of legs that transition from stance to swing as required in~\cite{fu2021minimizing}.

\subsection{Preliminaries}
Here we describe the low-level controllers to track the reference trajectories from the proposed energy-efficient planner.

\label{prelim}
\subsubsection{Swing Controller}
The swing controller is similar to~\cite{forceMPC}, which computes the torque for foot~$i$ for all three joints of the robot as\footnote{We omit time step $t$ for simplicity.}:
\begin{subequations}
\label{swing_controller}
\begin{align}
\boldsymbol{\tau}_{i}
=\ &
\mathbf{J}_{i}^{\top}\left[\mathbf{K}_{p}\left(\mathbf{p}_{i, \text {ref }} - \mathbf{p}_{i, \text  {cur }}\right)+\mathbf{K}_{d}\left(\mathbf{v}_{i, \text {ref }} -\mathbf{v}_{i, \text {cur }}\right)\right]
\\
&\
+\mathbf{J}_{i}^{\top} \boldsymbol{\Lambda}_{i} 
\mathbf{a}_{i, \text {ref}}
+\mathbf{V}_{i}(\dot{\mathbf{q}}_{i,\text{cur}}) 
+\mathbf{G}_{i}({\mathbf{q}}_{i,\text{cur}})
\end{align}
\end{subequations} 
where 
$\boldsymbol{\tau}_{i}\in \mathbb{R}^{3}$ is the joint torque, 
$\mathbf{q}_{i,\text{cur}} \in \mathbb{R}^{3 }$ and $\dot{\mathbf{q}}_{i,\text{cur}} \in \mathbb{R}^{3 }$ are the current joint position and velocity of foot $i$, 
$\mathbf{J}_{i}\in \mathbb{R}^{3 \times 3}$ is the foot Jacobian, 
$\mathbf{K}_{p}$ and $\mathbf{K}_{d}$ are the proportional and derivative (PD) gain matrices, 
$\mathbf{p}_{i, \text {ref}}\in \mathbb{R}^{3 }$ and $\mathbf{p}_{i, \text {cur}}\in \mathbb{R}^{3 }$ are the reference and current footstep positions (calculated using forward kinematics on current joint encoders) in the body frame, 
$\mathbf{v}_{i, \text {ref}}\in \mathbb{R}^{3 }$ and $\mathbf{v}_{i, \text {cur}} \in \mathbb{R}^{3 }$ are the reference and current footstep velocities in the body frame, 
$\mathbf{a}_{i, \text {ref}}\in \mathbb{R}^{3 }$ is the reference footstep acceleration in the body frame,
$\mathbf{V}_{i} \in \mathbb{R}^{3 }$ is the torque due to the Coriolis and centrifugal forces, 
$\mathbf{G}_{i} \in \mathbb{R}^{3 }$ is the torque due to gravity, 
and $\boldsymbol{\Lambda}_{i}\in \mathbb{R}^{3 \times 3}$ is the operational mass matrix.

\subsubsection{Stance Controller}
\label{sssec:stance}
For the legs in stance, i.e., making contact with the ground, we use a Model Predictive Controller (MPC) to calculate the ground reaction forces to track the reference trajectory of the body, or $\mathbf{X}_{\rm ref}\in \mathbb{R}^{12}$, where $\mathbf{X}_{\rm ref}=[\boldsymbol{\Theta}^\top,\mathbf{r}^\top,\boldsymbol{\omega}^\top,\mathbf{v}^\top]^\top$, 
$\boldsymbol{\Theta}$ is the robot's orientation, 
$\mathbf{r}$ is the CoM base position, 
$\boldsymbol{\omega}$ is the angular velocity, 
and $\mathbf{v}$ is the linear velocity, each vector with $x$, $y$, and $z$ components. 
While details of such controller are in \cite{forceMPC}, here, the ground reaction forces are used to calculate the joint torques that serve as input to the motor's torque controller:
\begin{equation}
\label{forceMPC}
\boldsymbol{\tau}_{i,t}=\mathbf{J}_{i,t}^{\top} \mathbf{R}^{w,\top}_{b,i,t} \mathbf{f}_{i,t}
\end{equation} 
where $\mathbf{f}_{i,t}\in\mathbb{R}^3$ is the force vector associated with leg~$i$ as subset of $\mathbf{f}_{t}$, and $\mathbf{R}^{w}_{b,i,t}$ is the rotation matrix from world to body frame of leg~$i$ at time step $t$.

\subsubsection{Reinforcement Learning through PPO}
\label{rl_ppo}
We employ Proximal Policy Optimization (PPO) \cite{schulman2017proximalpolicyoptimizationalgorithms} for both our optimal parameter study (Sec.~\ref{footstep_planner}), and for our model-free and motion imitation baselines. Only the reward formulation differs for each use case. In short, PPO enhances the policy $\pi_\theta(\mathbf{a} \mid \mathbf{O})$ by maximizing the clipped objective function:
\begin{equation}
L(\theta) = \mathbb{E}_t \left[ \min \left( r_t(\theta) \hat{A}_t, \text{clip}(r_t(\theta), 1 - \epsilon, 1 + \epsilon) \hat{A}_t \right) \right],
\end{equation}
Here, $r_t(\theta)$ denotes the ratio of the new policy to the old one, $\hat{A}_t$ represents the estimated advantage, and $\epsilon$ is a threshold that constrains policy updates to maintain stability. At each time step $t$, the PPO-derived policy outputs an action $\mathbf{a}_t$, which is then executed on the system: $\mathbf{a}_t = \pi_\theta(\mathbf{O}_t)$, where $\mathbf{O}_t$ refers to the current observation.

The observation space at time step $t$ for our model-free baseline is denoted by $\mathbf{O}_{t}^{M}$ while for our motion imitation baseline as well as for optimizing our planner parameter selection (see Sec.~\ref{footstep_planner}) is denoted by $\mathbf{O}_{t}^{I}$ (joints from all legs $i$ are considered in the observation space): 

\begin{equation}
\label{observation_combined}
\begin{aligned}
\mathbf{O}_{t}^{\rm M} &= [\mathbf{q}_{\text{cur}}, \dot{\mathbf{q}}_{\text{cur}}, \boldsymbol{\tau}, \mathbf{X}_{\text{cur}}, \mathbf{a}_{t-1}]_t, \\
\mathbf{O}_{t}^{I} &= [\mathbf{q}_{\text{cur}}, \dot{\mathbf{q}}_{\text{cur}}, \boldsymbol{\tau}, \mathbf{X}_{\text{cur}}, \mathbf{a}_{t-1}, \mathbf{q}_{\text{ref}}, \mathbf{X}_{\text{ref}}]_t,
\end{aligned}
\end{equation}
where the new variables include current body state described by $\mathbf{X}_{\text{cur}}$, the previous actions by $\mathbf{a}_{t-1}$, and reference joint trajectory by $\mathbf{q}_{\text{ref}}$. Note, although not shown explicitly in equation \eqref{observation_combined}, we do include the observational history up to $N_{\rm prev}$ as part of our overall observation space, or $\mathbf{O}_t=[\mathbf{O}_{t:t-N_{\rm prev}}]$. During model-free and motion imitation training, $\mathbf{a}_{t}$ represents desired joint positions, whereas in energy-efficient motion planning it serves as parameters of our planner such as swing time or ellipse shape (Sec. \ref{footstep_planner}).

Different combination of rewards were used depending on the RL training, although hyper-parameters were kept constant and $w$, $\sigma_{\rm track}$ represent weights on reward terms (see Table \ref{table:rewards}). For the motion imitation baseline, we used all the terms listed in (1), and (2) in Table \ref{table:rewards}. We used data from \cite{RoboImitationPeng20} as motion priors, which involve re-targeting mocap data from a real dog. For the model-free RL baseline we used all the terms listed in (2), and (3) in Table \ref{table:rewards}. Finally, for the parameter selection study described in detail in Sec. \ref{footstep_planner}, we used only the terms listed in (4) in Table \ref{table:rewards}. Note, we do not include all details of our hyper-parameter selection and domain randomization, as they follow similar structure as found in many other RL locomotion works such as \cite{Lee_2020}.

\section{Energy-Efficient  Motion Planner}
\label{footstep_planner}

\begin{figure}[t]
    \centering
    \includegraphics[width=0.99\columnwidth]{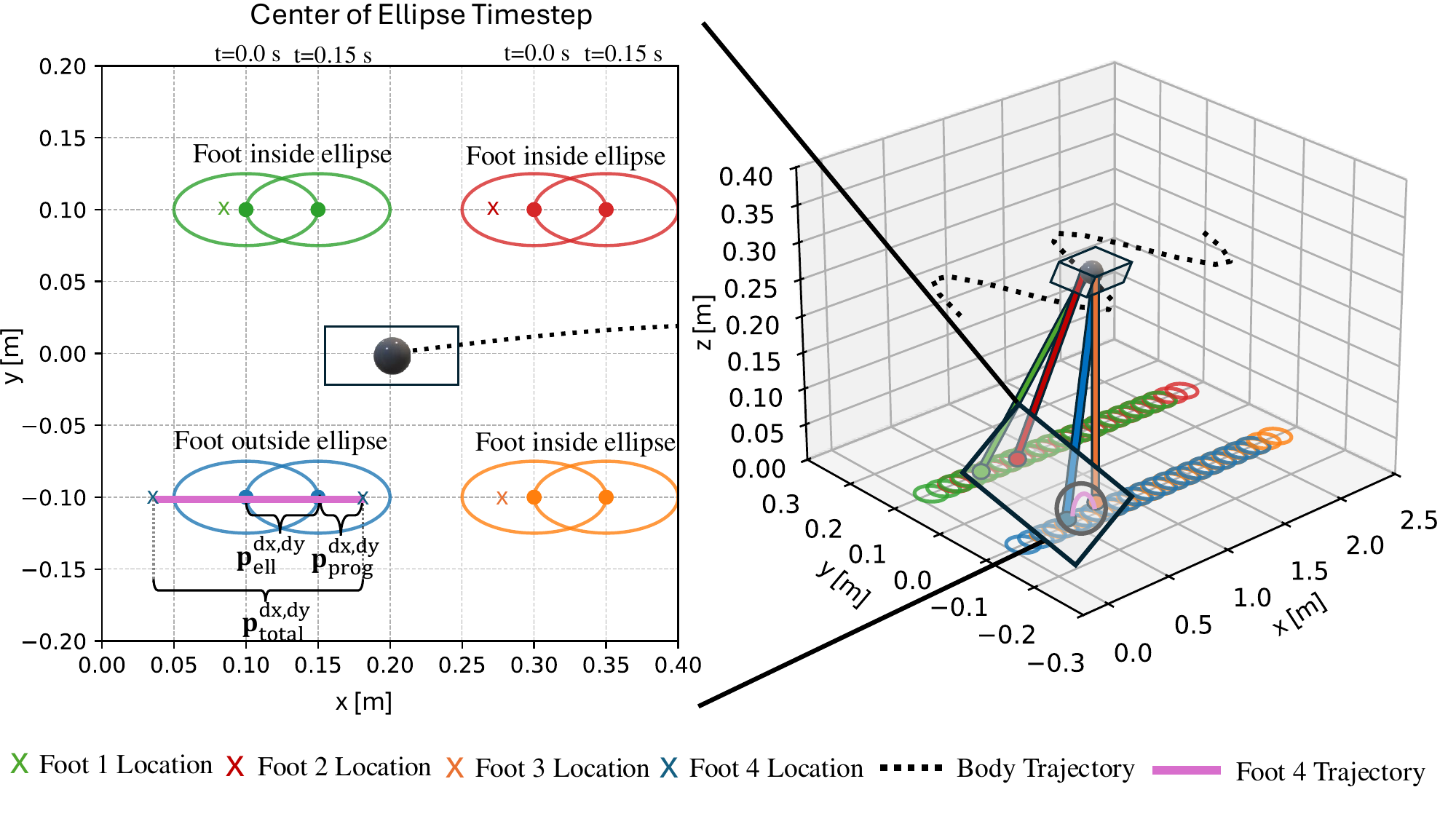}
    \caption{We show a 2D representation of the footstep ellipses and their variable representation as described in Sec. \ref{footstep_planner} on the left-hand side. A full 3D sequence is given in Fig. \ref{3d_sequence}.}
    \label{2d_sequence}
\end{figure}

Our planner is based on the relative position of the center of mass of the robot and the robot's feet.
In particular, it uses the geometric shape of an ellipse as the placement set for the feet. 
For each leg of the robot, the center point of the ellipse is always located directly beneath the hip joint, which imposes a stable position.
Other shapes for the placement sets can be used, but ellipses are an intuitive geometric shape for this purpose.
We define this ellipse through their major/minor axis, $r_{\rm ell}^{x}$ and $r_{\rm ell}^{y}$, with center point position, i.e., the robot's hip position, $p_{\rm hip}^{x}$ and $p_{\rm hip}^{y}$. 
As the robot's center of mass moves using a desired longitudinal and lateral commanded velocity, $\mathbf{V}_{\rm des}$, through the ground reaction forces calculated by the MPC~\cite{forceMPC} in Section~\ref{sssec:stance}, at each time step, we calculate the difference between the current footstep position, $p_{\rm cur}^{x}$ and $p_{\rm cur}^{y}$, and the hip position: 
\begin{subequations}
    \begin{align}
    p^{dx}_{\rm ell} 
    & = p_{\rm cur}^{x} - p_{\rm hip}^{x} 
    \\
    p^{dy}_{\rm ell} 
    & = p_{\rm cur}^{y} - p_{\rm hip}^{y}. 
\end{align}
\end{subequations} 

Given by a contact matrix $\mathbf{C}_{\rm ref}$ to indicate which feet are in stance or swing phase, using $p^{dy}_{\rm ell}$ and $p^{dy}_{\rm ell}$, we can check whether the current foot is outside this ellipse for each leg. For each foot in contact, the check can be represented by:
\begin{align} 
    \frac{(p^{dx}_{\rm ell})^2}{(r_{\rm ell}^{x})^2}  + \frac{(p^{dy}_{\rm ell})^2}{(r_{\rm ell}^{y})^2} > 1. 
    \label{ellipse_eqn}
\end{align}
If the check in~\eqref{ellipse_eqn} is true for any of the legs, we plan for the next footstep swing trajectory and corresponding body trajectory. The planner outputs the reference footstep position, velocity, and acceleration given by $\mathbf{p}_{\rm ref}, \mathbf{\dot{p}}_{\rm ref}, \mathbf{\ddot{p}}_{\rm ref}$ respectively, and the body trajectory, calculated using Euler discretization on the commanded velocity, $\mathbf{V}_{\rm des}$, and current state, $\mathbf{X}_{\rm cur}$, to yield the reference $\mathbf{X}_{\rm ref}$.

\begin{algorithm}[!ht]
    \SetAlgoLined
    \DontPrintSemicolon
    \BlankLine
    \While{Robot in Operation}{
        $\mathbf{X}_{\text{cur}}, \mathbf{p}_{\text{cur}} \leftarrow \text{StateEstimation}(\text{See \cite{state_estimator}})$ \\
        \tcp{Get user commands (10 Hz)}
        $\mathbf{V}_{\rm des} \leftarrow \text{Joystick}$ \\
        \tcp{Get planner parameters}
        $[\Delta T, h_{\rm step}, X^{z}, r_{\rm ell}^{x}, r_{\rm ell}^{y}]\leftarrow\text{Analytical or RL } \text{(Sec. \ref{footstep_planner})}$
        
        \BlankLine
        \tcp{For each leg}
        $\text{p}_{\rm ell}^{dx} = \text{p}_{\rm cur}^{x} - \text{p}_{\rm hip}^{x}$ \\
        $\text{p}_{\rm ell}^{dy} = \text{p}_{\rm cur}^{y} - \text{p}_{\rm hip}^{y}$ \\
        \BlankLine
        \tcp{Check if foot in contact is outside ellipse}
        $\text{Outside Ellipse?} \leftarrow \text{checkEllipse}(\text{p}_{\rm ell}^{dx}, \text{p}_{ell}^{dy}, \text{r}_{\rm ell}^{x}, \text{r}_{\rm ell}^{y})$ \\
        \BlankLine
        \tcp{If outside the ellipse, calculate a new reference trajectory}
        $\mathbf{if} \text{  Outside Ellipse} = \text{True}:$ \\
        \Indp
            $\text{p}_{\rm prog}^{dx}=V_{x,\rm des}^{B}\Delta T + \sqrt{X^{z}/g}(V_{x,\rm cur}^{B} - V_{x,\rm des}^{B})$ \\
            $\text{p}_{\rm prog}^{dy}=V_{y,\rm des}^{B}\Delta T + \sqrt{X^{z}/g}(V_{y,cur}^{B} - V_{y,\rm des}^{B})$ \\
            $\mathbf{p}_{\rm ref}, \mathbf{\dot{p}}_{\rm ref}, \mathbf{\ddot{p}}_{\rm ref}, \mathbf{C}_{\rm ref} \leftarrow \text{Sinusoidal}(dt, \Delta T, gait, \mathbf{p}_{\rm prog}^{dx,dy}, \mathbf{p}_{\rm ell}^{dx,dy}, h_{\rm step})$\\
            $\mathbf{X}_{\rm ref} \leftarrow \text{interp}(\mathbf{X}_{\rm cur},\mathbf{V}_{\rm des})$\\
        \Indm
        \tcp{If no legs are outside the ellipse nor currently in swing phase, we put all legs in stance, and include the reference commanded body velocity to push body forwards}
        $\mathbf{if} \text{ Swing Phase Complete} = \text{True}:$ \\
        \Indp
            $\mathbf{X}_{\rm next}, \mathbf{p}_{\rm next}, \mathbf{\dot{p}}_{\rm next}, \mathbf{\ddot{p}}_{\rm next}, \mathbf{C}_{\rm next} \leftarrow \mathbf{X}_{\text{cur}},  \mathbf{p}_{\text{cur}}, \mathbf{0}^{12}, \mathbf{0}^{12}, \mathbf{1}^{4} $ \\
        \Indm
        \tcp{Otherwise, iterate through the reference trajectory}
        $\textbf{else:}$ \\
        \Indp
            $\mathbf{X}_{\rm next}, \mathbf{p}_{\rm next}, \mathbf{\dot{p}}_{\rm next}, \mathbf{\ddot{p}}_{\rm next}, \mathbf{C}_{\rm next} \leftarrow \mathbf{X}_{\rm ref}[it], \mathbf{p}_{\rm ref}[it], \mathbf{\dot{p}}_{\rm ref}[it], \mathbf{\ddot{p}}_{\rm ref}[it], \mathbf{C}_{\rm ref}[it]$ \\
        \Indm
        Send references to swing \eqref{swing_controller} and stance \eqref{forceMPC} controllers.
    }
    \caption{Energy-Efficient Motion Planner}
    \label{planner}
\end{algorithm}

\begin{table}[h]
\caption{Reward functions. See Sec. \ref{rl_ppo} for Details on Usage.}
\centering
\renewcommand{\arraystretch}{1.5}
\begin{tabular}{p{4cm}|p{3.3cm}}
\arrayrulecolor[HTML]{FF0000}\hline
\rowcolor[HTML]{FFFFFF} \textbf{(1) Term (Imitation)} & \textbf{Equation} \\
\arrayrulecolor[HTML]{FF0000}\hline
\rowcolor[HTML]{F7F7F7} Joint Position Reward & $\left(e^{-\frac{\sum \|\mathbf{q}_{\text{cur}} - \mathbf{q}_{\text{ref}}\|^2}{\sigma_{\rm track}}}\right)w_1$ \\
\rowcolor[HTML]{FFFFFF} Foot Position Reward & $\left(e^{-\frac{\sum \|\mathbf{p}_{\text{cur}} - \mathbf{p}_{\text{ref}}\|^2}{\sigma_{\rm track}}}\right)w_2$ \\
\arrayrulecolor[HTML]{0000FF}\hline
\rowcolor[HTML]{FFFFFF} \textbf{(2) Term (Tracking)} & \textbf{Equation} \\
\arrayrulecolor[HTML]{0000FF}\hline
\rowcolor[HTML]{F7F7F7} Linear Velocity in XY Plane & $\left(e^{-\frac{\|\mathbf{v}_{\text{cur}, \text{xy}} - \mathbf{v}_{\text{ref}, \text{xy}}\|^2}{\sigma_{\rm track}}}\right)w_3$ \\
\rowcolor[HTML]{FFFFFF} Angular Velocity in Z Axis & $\left(e^{-\frac{\|\omega_{\text{cur}, z} - \mathbf{\omega}_{\text{ref}, z}\|^2}{\sigma_{\rm track}}}\right)w_4$ \\
\rowcolor[HTML]{F7F7F7} Linear Velocity in Z Axis & $-\left(\|\mathbf{v}_{\text{cur}, z} - \mathbf{v}_{\text{ref}, z}\|^2\right)w_5$ \\
\arrayrulecolor[HTML]{00AA00}\hline
\rowcolor[HTML]{FFFFFF} \textbf{(3) Term (Regularization)} & \textbf{Equation} \\
\arrayrulecolor[HTML]{00AA00}\hline
\rowcolor[HTML]{F7F7F7} Joint Accelerations & $-\left(\sum \|\dot{\mathbf{q}}\|^2\right)w_{6}$ \\
\rowcolor[HTML]{FFFFFF} Action Rate & $-\left(\sum \|\mathbf{a}_t - \mathbf{a}_{t-1}\|^2\right)w_{7}$ \\
\rowcolor[HTML]{F7F7F7} Foot Airtime & $\left(\sum t_{\text{air}} 1_{\text{new\_contact}}\right)w_{8}$ \cite{Lee_2020} \\
\rowcolor[HTML]{FFFFFF} Self-collision & $\left(1_{\text{collision}}\right)w_{9}$ \cite{Lee_2020} \\
\arrayrulecolor[HTML]{FFA500}\hline
\rowcolor[HTML]{FFFFFF} \textbf{(4) Term (Parameter Selection)} & \textbf{Equation} \\
\arrayrulecolor[HTML]{FFA500}\hline
\rowcolor[HTML]{F7F7F7} Cost of Transport & $-\left(\text{CoT}\right)w_{10}$ eq.\eqref{cost_of_transport} \\
\rowcolor[HTML]{FFFFFF} Robustness & $\left(\text{Manipulability}\right)w_{11}$ eq.\eqref{eq:manipulability} \\
\rowcolor[HTML]{F7F7F7} Number of Time Steps & $\left(t/t_{\rm max}\right)w_{12}$ \\
\end{tabular}
\label{table:rewards}
\end{table}

When all feet are inside their corresponding ellipse, the reference trajectory stays in place. 
In other words, if the user commands a forward velocity, the MPC controller pushes the robot's body in that direction with all feet in stance. Once one of the foot locations is outside the ellipse, the planner produce a swing trajectory. 
The reference trajectory for the footstep in swing is calculated by moving the current footstep position first to the center of the ellipse, and then adding an additional term, $p_{\rm prog}^{dx}$ and $p_{\rm prog}^{dy}$ (see Fig. \ref{2d_sequence}), to account for the desired commanded velocity:
\begin{subequations}
\label{eq:progress}
\begin{align}
    p^{dx}_{\rm prog} 
    & = V_{x,\rm des}^{B}\Delta T/2 + \sqrt{X^{z}/g}(V_{x,\rm cur}^{B} - V_{x,\rm des}^{B}) 
    \\
    p^{dy}_{\rm prog} 
    & = V_{y,\rm des}^{B}\Delta T/2 + \sqrt{X^{z}/g}(V_{y,\rm cur}^{B} - V_{y,\rm des}^{B})
\end{align}
\end{subequations}
where $X^{z}$ is the height of the robot, $g$ is the gravity term, $\Delta T$ is the footstep timing for the swing phase, and $B$ is to symbolize that we expect the desired velocity, $V_{x,\rm des}^{B}$ and $V_{y,\rm des}^{B}$, to be in the body frame. Note, this is derived from the Raibert Heuristic \cite{Raibert}, where the the feedback gain $k$ is chosen to be equal to $\sqrt{X^{z}/g}$.

\newcommand{\steptime}{\Delta T}
The reference trajectory from the current footstep $\mathbf{p}_{\rm cur}$ to the next footstep or $\mathbf{p}_{\rm next}$ is created using a sinusoidal implementation:
\begin{subequations}
\label{eq:sinus}
\begin{align}
\bar a_{z} &= h_{\rm step} \frac{1}{2} \left(\frac{2\pi}{\steptime}\right)^2 
\\
a_{z,t} &= \bar a_{z} \cos \left(\frac{2\pi}{\steptime} t\right)
\\
v_{z,t} &= \bar a_{z} \frac{\steptime}{2\pi} \sin \left(\frac{2\pi}{\steptime} t\right)
\\
p_{z,t} &= 
\bar a_{z} \left(\frac{\steptime}{2\pi}\right)^2
\left(1 - \cos \left(\frac{2\pi}{\steptime} t\right)\right)
\end{align} 
with step height $h_{\rm step}$, maximum vertical acceleration $\bar a_{z}$, and:
\begin{align}
\bar a_{x} &= 
\left(
p^{dx}_{\rm ell} 
 +
p^{dx}_{\rm prog}  
\right)
\frac{1}{2}  \left(\frac{\pi}{\steptime}\right)^2
\\
\bar a_{y} &= 
\left(
p^{dy}_{\rm ell} 
 +
p^{dy}_{\rm prog}  
\right)
\frac{1}{2}  \left(\frac{\pi}{\steptime}\right)^2
\\
a_{x/y,t} &= \bar a_{x/y} \cos \left(\frac{\pi}{\steptime} t\right)
\\
v_{x/y,t} &= \bar a_{x/y} \frac{\steptime}{\pi} \sin \left(\frac{\pi}{\steptime} t\right)
\\
p_{x/y,t} &= \bar a_{x/y} \left(\frac{\steptime}{\pi}\right)^2
\left(
1
- 
\cos \left(\frac{\pi}{\steptime} t\right)
\right)
\end{align} 
\end{subequations}
with maximum lateral acceleration $\bar a_{x/y}$.
Note that the proposed footstep motion planner does not rely on the choice of trajectory. 
Different implementations can be used as well. 
In particular, for climbing stairs or traversing rubble and obstacles,~\eqref{eq:sinus} may be modified to alter the touchdown point.

The contact matrix $\mathbf{C}_{\rm ref} \in \mathbb{R}^{4}$ specifying which leg is in stance or swing phase can be pre-determined if a certain gait is desired, e.g., trot or walk, or can be determined by the algorithm. 
E.g., in the pre-determined case for a trot gait, if one of the footsteps is outside the defined ellipse, the foot along with the diagonally opposing footstep are constrained to move at the same time from stance into swing. 
In this case,  we only move the leg and its diagonally opposing leg in alternating order, e.g., if legs 1 and 3 were in swing previously, we must move legs 2 and 4 next. 
In the non-predetermined case, we remove such constraints (except that we cannot move the same leg twice). 
Instead, we only move the leg that passes~\eqref{ellipse_eqn}, allowing $\mathbf{C}_{\rm ref}$ to freely vary. In both cases, since foot movement follows~\eqref{ellipse_eqn}, stance phase duration varies with current body position relative to the feet. Thus, unique gaits and footstep timings for each step are generated.

We now detail how to enforce energy-efficiency for our motion planner. A common metric for energy-efficient planning is the Cost of Transport (CoT), which can be computed using the joint velocity $\dot{\mathbf{q}}_{t}$, torque $\boldsymbol{\tau}_{t}$, and distance $\Delta D$ traveled:
\begin{equation}
\label{cost_of_transport}
\begin{aligned}  
\text{\rm CoT}
= 
\frac{\sum_{t=k-N}^{k} 
\max \{\dot{\mathbf{q}}_{t}^\top \boldsymbol{\tau}_{t} , 0\}
{\rm d}t}{\Delta D} 
\end{aligned} 
\end{equation}
where $k$ is the current time step discretized with sampling time ${\rm d}t$, and $N$ is the number of time steps within total time of locomotion considered.

For each desired velocity commanded by the user ($\mathbf{V}_{\rm des}$), we determine ideal parameters for the proposed planner that achieve not only the lowest CoT but also favors robot configurations for which the legs in stance can most easily produce ground reaction forces. 
To quantify the favorable configurations, we leverage force manipulability ellipsoids~\cite{MELCHIORRI1994235}. 
These ellipsoids provide a concise and intuitive representation of the force and moment capabilities of a robot at a particular point in its workspace. Mathematically, the motor torque vector is mapped into a unit sphere $\boldsymbol{\tau}_{i}^{\top}\boldsymbol{\tau}_{i} \leq \mathbf{1}$ in joint space and into the ellipsoids $\mathbf{f}^\top_{i} \mathbf{J}_{i}\mathbf{J}_{i}^{\top}\mathbf{f}_{i} \leq \mathbf{1}$ in task space, where $\mathbf{J}_{i}$ and $\mathbf{f}_{i}$ are the Jacobian and end-effector ground reaction forces, respectively. 
We can estimate a representation of the volume of these force ellipsoids through the force manipulability measure~\cite{MELCHIORRI1994235}. 
A high force manipulability measure indicates a better ability to apply forces in various directions, while a lower manipulability measure indicates a more limited ability to do so. To calculate this force manipulablity measure, we first take the eigenvalues, $\boldsymbol{\lambda}$, of the matrix $U=(\mathbf{J}_{i}\mathbf{J}_{i}^{\top})^{-1}$, and: 
\begin{equation}
\label{eq:manipulability}
\begin{aligned}  
\text{manipulability measure} = \sqrt{\max(\boldsymbol{\lambda})/\min(\boldsymbol{\lambda})}.
\end{aligned} 
\end{equation}

Thus, our goal is to find the parameters for each commanded body velocity that minimize the CoT and maximize the manipulability measure of the force ellipsoids for increased robustness. 
To do so, we compute both CoT and the force manipulability measure as a function of the planner parameters using Mujoco \cite{todorov2012mujoco} simulation as our physics model for propagating from the current to the next time step. 
In particular, we study the impact of the swing time $\Delta T$, body velocity in the $x$-direction $V_{x,\rm des}^{B}$, step height $h_{\rm step}$, robot height $X^{z}$, and the ellipse size of the planner $r_{\rm ell}^{x}$ and $r_{\rm ell}^{y}$. 
For each combination of parameters, we calculate the average CoT and force manipulability measure during about 20 seconds of locomotion. 
The range of parameters tested is described in Table~\ref{tab:analytical}. 
To calculate the CoT and the force manipulablity measure, the joint torques for the feet in swing and stance are needed. These torques can be received directly from Mujoco. 
We call this method of parameter selection as Energy-Efficient Motion Planner using the \textit{analytical approach} (or EEMP-Analytical), because we can find these parameters without any learning methods---we simply iterate through a range of velocities to find which parameter combination produced the ideal CoT and manipulability measure for each commanded velocity and employ a look-up table during evaluation.

In addition to this \textit{analytical approach}, we also employ a simple RL formulation using PPO (Sec.~\ref{rl_ppo}) to find these parameters through exploration and compare against the \textit{analytical approach}. We term this as EEMP-RL. In this approach, the actions are the parameters of our planner from Table~\ref{tab:analytical}, or 
$\mathbf{a}_{t} = [\Delta T, h_{\rm step}, X^{z}, r_{\rm ell}^{x}, r_{\rm ell}^{y}]_t$, and the observations are described by $\mathbf{O}_t^{I}$ in equation~\eqref{observation_combined}, using IK on $\mathbf{p}_{\text{ref}}$ to get $\mathbf{q}_{\text{ref}}$. Our reward function consist of only the three terms listed in (4) of Table \ref{table:rewards}, which is to penalize the agent for high CoT, reward high manipulability, and to reward the agent for not falling to the ground---where $t$ is the current time step during an episode, and $t_{\rm max}$ is the maximum allotted time step per episode or 20 seconds of locomotion.

\begin{table}[]
    \centering
        \caption{Parameters Used for Parameter Study}
\begin{tabular}{|c|c|c|c|c|}
\toprule Name & Description & min & max & units \\
\midrule $V_{x,\rm des}^{B}$ & Desired Body Velocity & 0.05 & 1.00 & $\frac{\rm m}{\rm s}$ \\
$\Delta T$ & Swing Time & 0.10 & 0.25 & s \\
$h_{\rm step}$ & Step height & 0.05 & 0.15 & m \\
$X^{z}$ & Robot Height & 0.28 & 0.311 & m \\
$r_{\rm ell}^{x}$, $r_{\rm ell}^{y}$ & Ellipse axis in x/y direction & 0.01 & 0.15 & m \\
\bottomrule
\end{tabular}
\label{tab:analytical}
\end{table}

The overall footstep planner is described in Algorithm~\ref{planner}, and visually shown in 3D and 2D in Figs. \ref{3d_sequence} and \ref{2d_sequence} respectively.
Summarizing, we first get the state of the robot \cite{state_estimator}, and then receive the desired velocity command from the user. Planner parameters are then selected through either EEMP-Analytical or EEMP-RL.
We then 
check~\eqref{ellipse_eqn} to see if any of the footsteps in contact with the ground are currently outside the defined ellipse. 
If one of the footsteps is outside the ellipse, we calculate the footstep trajectory~\eqref{eq:progress}, \eqref{eq:sinus} and the body trajectory through Euler discretization for the commanded velocity.   
Additionally, we check if we fully iterated through the swing trajectory based on the swing time $\Delta T$. 
If so, we produce four stance phase reference trajectories for all legs, i.e., the robot feet are commanded to stay in position, while the body reference, $\mathbf{X}_{\rm next} \in \mathbb{R}^{12}$, is such that it moves in the direction of the desired velocity. 
If the swing phase is not completed, we proceed to the next time step of the swing reference trajectory. References are then sent to our downstream stance and swing controllers.

\section{Parameter Study Results}
We first applied the EEMP-Analytical approach (Sec.~\ref{footstep_planner}) to examine how optimal parameter selection affects CoT and manipulability across various lateral velocities ($x$-direction) during 20 seconds of locomotion and setting the gait to be either in walk or trot to compare both. Our analysis showed that the parameters in Table~\ref{tab:baseline} consistently yielded the lowest CoT and highest manipulability \textit{on average}, i.e., across the velocity range in walk/trot. However, optimizing parameters for each velocity  improved CoT by 35.5\% in the trot gait and 13.3\% in the walk gait over using the constant set of parameters from Table~\ref{tab:baseline}.

\begin{table}[t]
    \centering
        \caption{Best Parameters for Walk / Trot Gait Across Velocity Range}
\begin{tabular}{|c|c|c|c|}
\toprule Name & Description & Value & units \\
\midrule
$\Delta T$ & Swing Time & 0.25 & s \\
$h_{\rm step}$ & Step height & 0.10 & m \\
$X^{z}$ & Robot Height & 0.31 & m \\
$r_{\rm ell}^{x}$, $r_{\rm ell}^{y}$ & Ellipse axis in x/y directions & [0.07, 0.05] & m \\
\bottomrule
\end{tabular}
\label{tab:baseline}
\end{table}

\begin{figure}[!t]
    \centering
    \includegraphics[width=1.0\columnwidth]{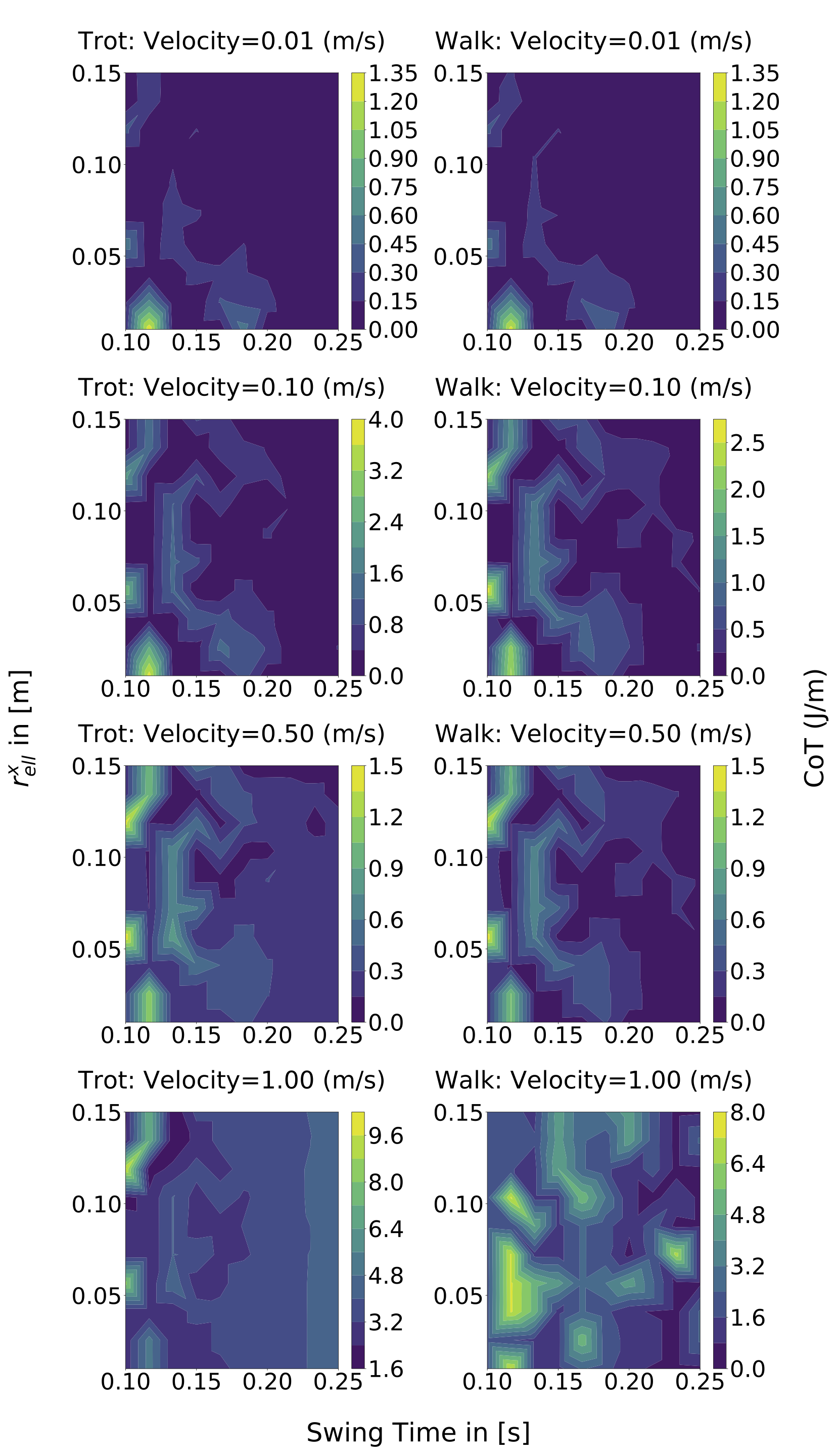}
    \caption{Contour plots for different commanded velocities (namely, 0.01, 0.10, 0.50, and 1.00 m/s), and for varying major axis of the planning ellipse $r_{\rm ell}^{x}$ (y-axis) and swing time (x-axis). 
    The color varies with the values of CoT, from yellow to blue. 
    The graphs on the left half show the trot gait results, while the graphs on the right half show the walk gait results.} 
    \label{fig:contour}
\end{figure}

Fig.~\ref{fig:contour} shows contour plots of the parameter selection,
where for each commanded velocity we show the ellipse's major $x$ axis, $r^{x}_{\rm ell}$ (toward the locomotion direction), swing time, and color coding represented by CoT. 
The results of the trot gait are shown on the left and of the walk gait on the right half.
For the trot gait, we show that high swing times are critical at low velocities, whereas the selection for $r^{x}_{\rm ell}$ appears less important. As velocity increases to 0.50 m/s, the combination of $r^{x}_{\rm ell}$ and swing time becomes critical, where a swing time of about 0.18 seconds and $r^{x}_{\rm ell}$ of about 0.06 to 0.08 meters, or high swing time and high $r^{x}_{\rm ell}$ leads to optimal CoT. However, at velocities going towards 1.0 m/s, lower swing times of 0.14-0.16~seconds become more favorable, while $r^{x}_{\rm ell}$ has minimal impact in this range. This is intuitive since, at higher velocities, the footstep often falls outside the ellipse, triggering constant step initiation regardless of $r^{x}_{\rm ell}$.

\begin{figure}[t]
    \centering
    \includegraphics[width=0.99\columnwidth]{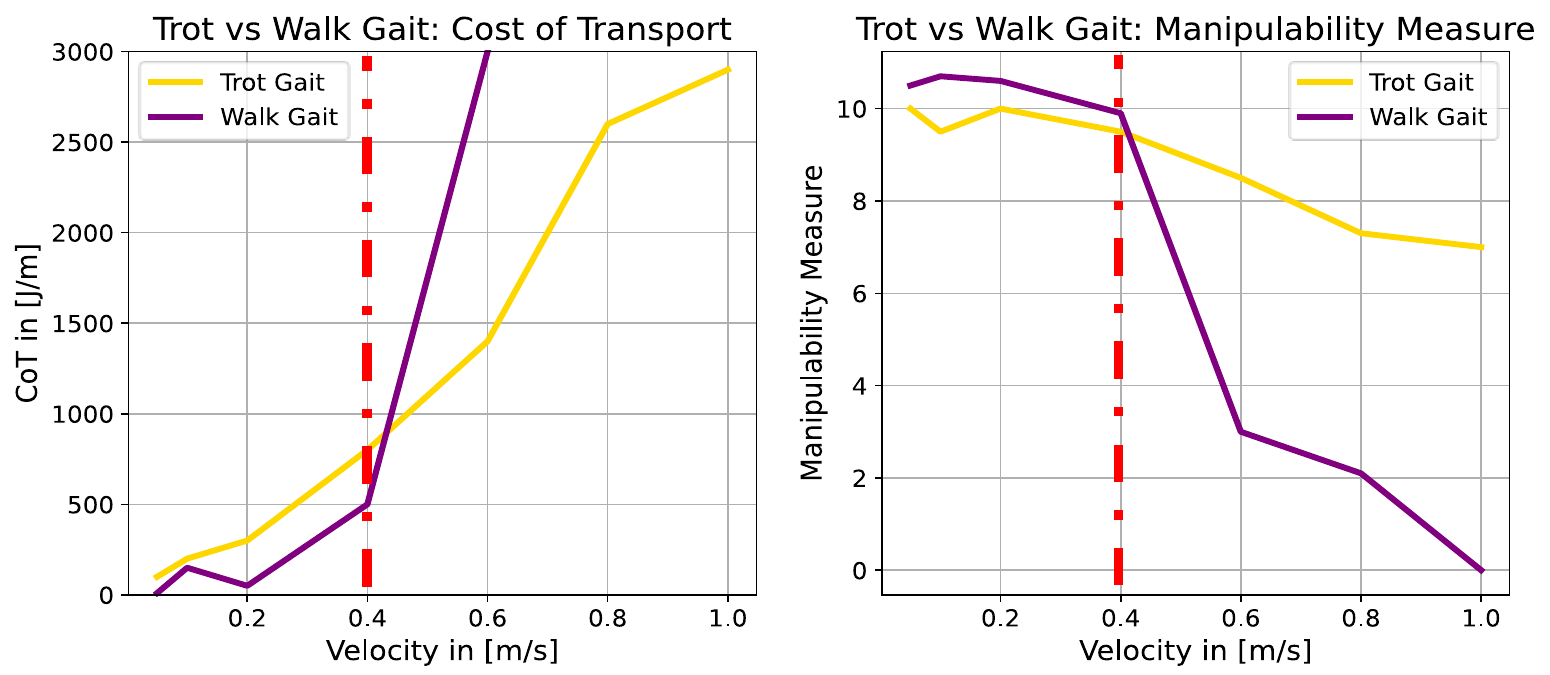} 
    \caption{CoT (left) and the manipulability measure (right) for the trot (yellow) and walk (purple) gaits at various velocities. The red line indicates the velocity at which walk gait becomes unstable.
    }
    \label{fig:walk_vs_trot}
\end{figure}

For the walk gait, we see the same trend as for the trot gait in low velocities. However, different from the trot gait, a high swing time is more favorable at nearly all choice of velocities. Although the $r^{x}_{\rm ell}$ parameter seems more important if the user selects lower swing times, at higher velocities $r^{x}_{\rm ell}$ loses effect, particularly at 1.0 m/s due to a similar reason as for the trot gait. We make a comparison for the CoT and manipulability measure between the trot and walk gait in Fig.~\ref{fig:walk_vs_trot}. In the velocity range up to 0.4 m/s, the walk gait has both a more favorable CoT and manipulability measure relative to trot. This is expected as energy consumption is estimated by summing joint accelerations and torques across all legs, with the swing phase being most significant. Since walk gait has one leg in swing phase at a time, it results in lower energy consumption across all velocities, as confirmed by our results. As the manipulability measure is based on how well the robot can produce ground reaction forces at any given time, the walk gait is expected to show higher performance as well since it has three legs in stance phase, instead of two legs for the trot gait. Overall, the walk gait used 59\% of energy of the trot gait in this velocity range. At velocities exceeding 0.4 m/s (marked by the red line in Fig. \ref{fig:walk_vs_trot}), the walk gait becomes unstable, evidenced by a sharp increase in CoT and a decline in the manipulability measure. This instability is consistently observed in both simulation and hardware experiments at velocities above 0.4 m/s.



\begin{figure}[t]
    \centering
    \includegraphics[width=0.99\columnwidth]{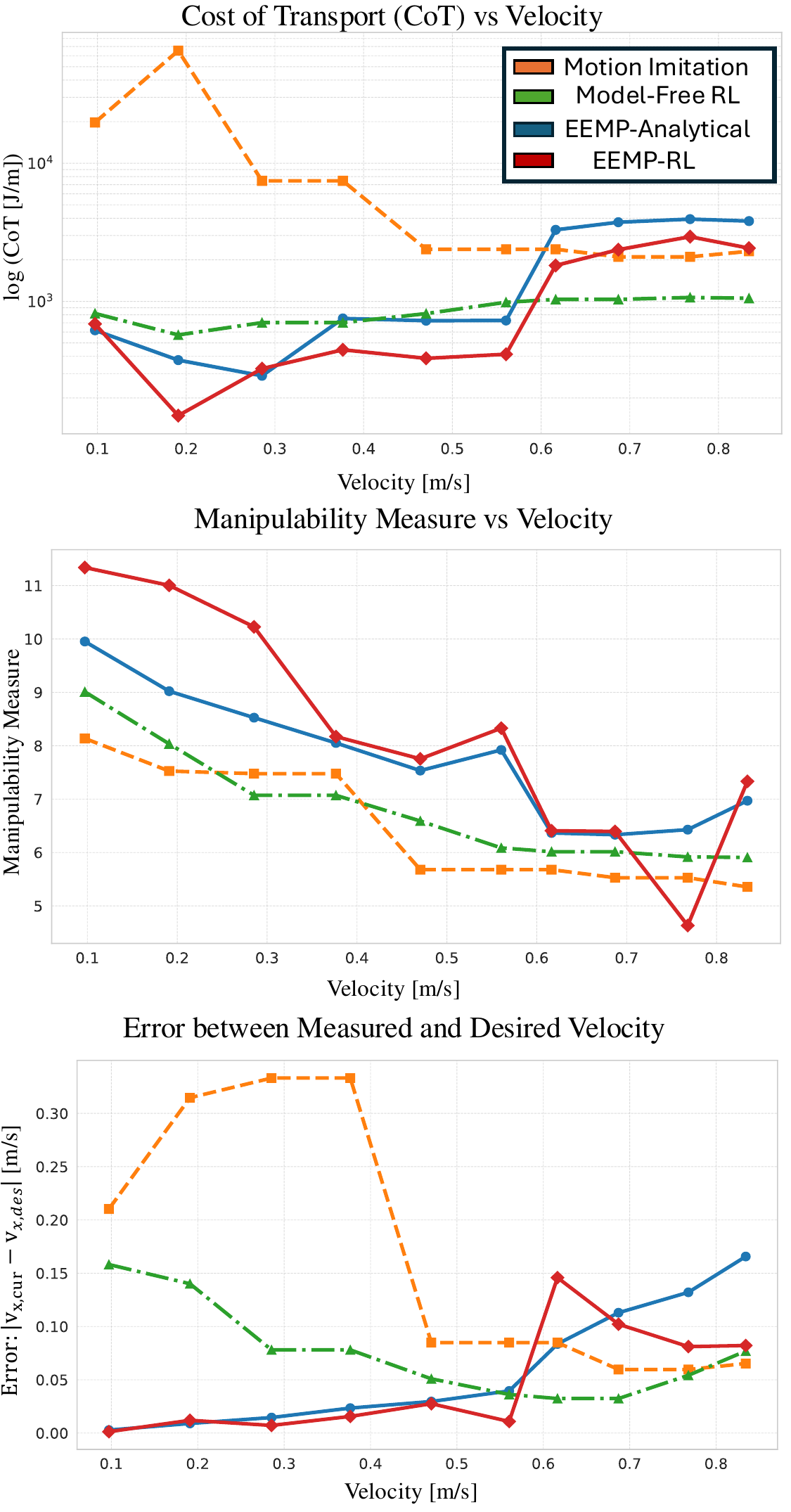}
    \caption{CoT, manipulability measure, and velocity tracking (y-axis) for various desired velocities (x-axis) are shown, comparing our Energy-Efficient Motion Planner (EEMP) using the analytical approach and RL approach for parameter selection in blue and red respectively, motion imitation in orange, and model-free RL in green.} 
    \label{fig:baseline_vs_optimal}
\end{figure}

Finally, we also compare EEMP-Analytical with parameter selection using RL (EEMP-RL, Sec.~\ref{footstep_planner}). As shown in Fig.~\ref{fig:baseline_vs_optimal}, EEMP-RL (red) slightly outperforms EEMP-Analytical (blue) in both CoT and manipulability. This indicates two key points: first, our method achieves efficient locomotion without requiring RL, as both methods show good performance; second, RL improves performance by better handling disturbances from estimation errors, contacts, or velocity transitions. For instance, RL can optimize parameters in continuous space for smoother gait changes, instead of relying on a discretized look-up table (i.e., EEMP-Analytical), thereby enhancing overall energy efficiency.


\begin{figure}[t]
    \centering
    \includegraphics[width=0.99\columnwidth]{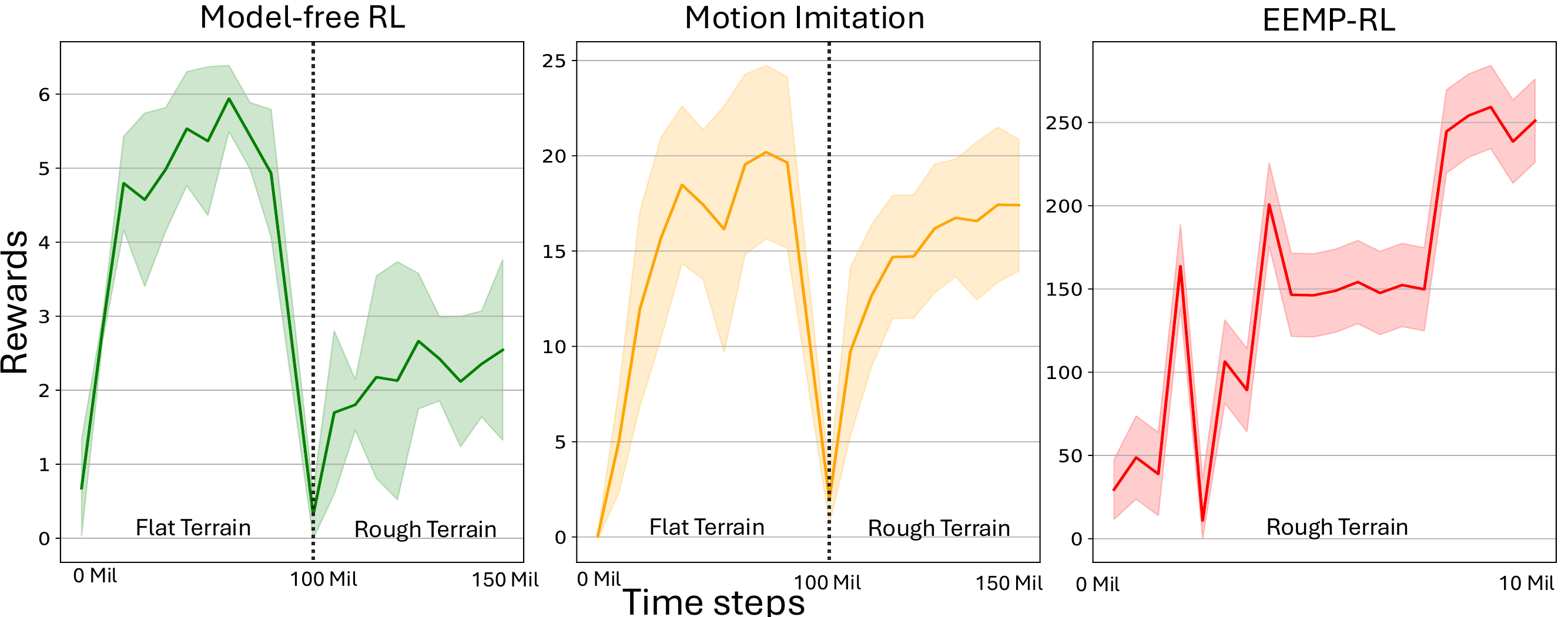}
    \caption{Rewards for model-free RL and motion imitation baselines, and our Energy-Efficient Motion Planner using RL (EEMP-RL).} 
    \label{fig:training_rewards}
\end{figure}

\section{Experimental Validation}

\subsubsection{Controller Architecture and Baseline Comparisons}
\label{controller_architecture}
In simulation and hardware, the velocity commands are provided by the operator, via a joystick. For state estimation we use a similar approach as \cite{state_estimator}. (A) in Fig.~\ref{fig:robot} summarizes the control architecture and control frequency used in this paper for both Mujoco simulation and hardware experiments. We also show two other baselines methods for achieving legged locomotion, namely through motion imitation and model-free RL (Sec. \ref{rl_ppo}). The results of training these baselines, along with our optimal parameter selection (EEMP-RL) is shown in Fig.~\ref{fig:training_rewards}. We note that we trained our RL baselines on flat terrain for 100 million time steps, and on rough terrain for 50 million time steps (thus, the drop in rewards occur due to the switch of terrains, which is expected), and we required only 10 million time steps directly on rough terrain for EEMP-RL.


\subsubsection{Mujoco Simulation Results}
For 20 seconds of locomotion per desired velocity, we compare our method, EEMP-Analytical (blue) and EEMP-RL (red), without predefining walk/trot for parity, with motion imitation (orange) and model-free RL (green) baselines using CoT, manipulability, and velocity tracking as evaluation metrics. Overall, our method performs best at $\leq$ 0.6 m/s across all metrics (i.e., CoT, manipulability, and velocity tracking). At higher velocities, model-free RL and motion imitation slightly outperform EEMP methods in CoT, but perform similarly in velocity tracking and manipulability using EEMP-RL. This is expected, as higher velocities shrink the ellipses to a point, requiring constant stepping, thus reducing the energy-efficiency benefits of our planner. Additionally, nonlinear dynamics at high speeds challenge the linear MPC stance controller assumption underlying the EEMP framework, a limitation of this work. Addressing this may require nonlinear models, at the cost of higher computation, or learning such models offline with neural nets. Notably, model-free RL benefits from tightly coupling the planner and controller (unlike our EEMP methods), which proves advantageous at higher speeds. Interestingly, motion imitation under performs compared to model-free RL at nearly all velocities, suggesting that biological imitation may be less effective for lower-DoF robots, such as a 3-DoF per leg quadruped. Based on these findings, the most optimal energy-efficient motion would be one that uses EEMP-RL for velocities $\leq0.6$ m/s, and then transition to model-free RL.

\subsubsection{Hardware Validation}

\begin{figure}[t]
    \centering
    \includegraphics[width=0.99\columnwidth]{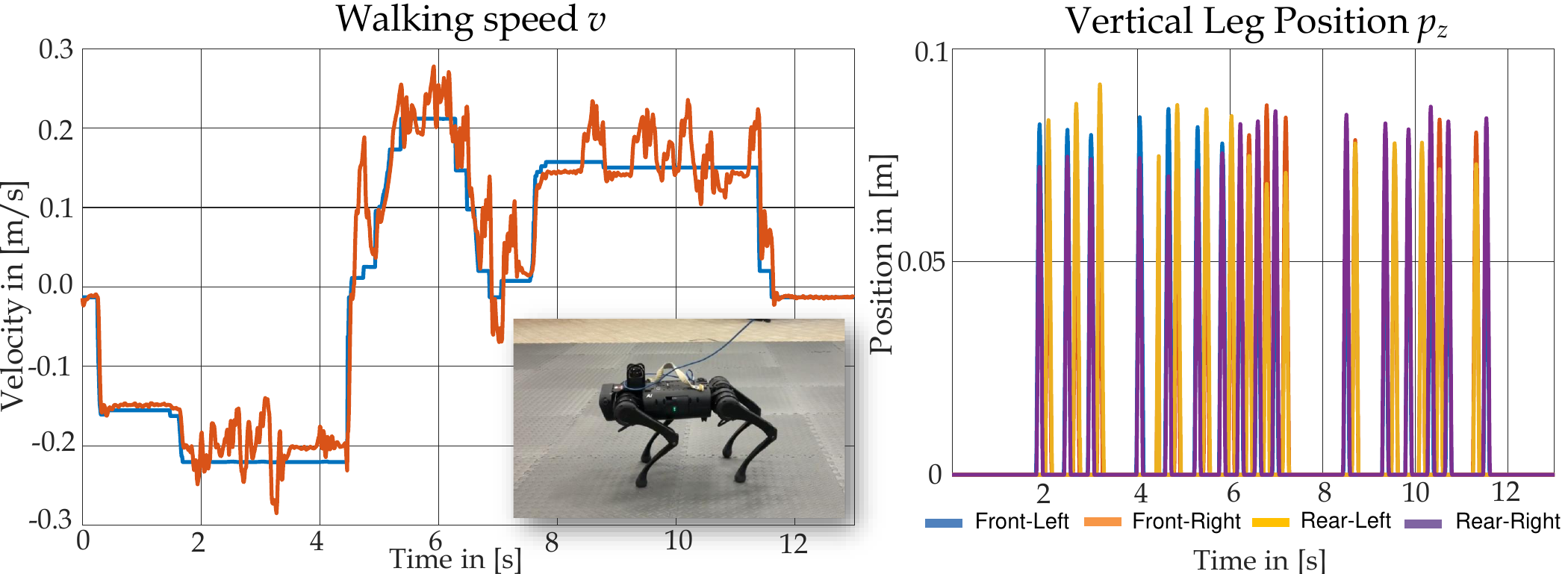} 
    \caption{Results of teleoperation with our architecture to demonstrate robustness against disturbances on hardware. Using a joystick, we commanded the robot with velocities shown in blue arbitrarily selected, while orange indicates the actual velocity received from state estimation (left graph). The corresponding vertical leg positions are shown in the right graph. 
    }
    \label{fig:footstep_experiments}
\end{figure}


We teleoperated the robot shown in Fig.~\ref{fig:footstep_experiments} using a desired commanded walking speed to test how the proposed footstep planner and control framework handles disturbance rejection. 
Fig.~\ref{fig:footstep_experiments} shows an example of a trot gait, where the left plot shows the commanded forward velocity and the robot's velocity from the state estimator. 
The vertical leg trajectories are shown on the right figure. 
From about 6.5 to 8.5~s, the vertical positions of the legs are zero indicating that all four legs are in stance. The left plot shows a forward walking speed in this period. 
This is a feature of the planner, the robot only executes a step when necessary, specifically, when the foot location is outside the specified ellipse. 
Thus, the body tracks the commanded velocity while conserving energy for some seconds by pushing the stance legs to move the body forward without taking a step.

\section{Conclusions}
A motion planner for legged robots was developed that is energy efficient. We show a 34.5\% improvement in CoT over a baseline with fixed parameters for trot gait, and a 13.3\% improvement in walk gait by employing a parameter design framework that uses a motion model based on Newtonian mechanics. Overall, we found that the walk gait used 59\% of energy of the trot gait. For low to medium velocity, our method outperforms model-free RL, in CoT, manipulability, and velocity tracking. We show a 50.4\% improvement in CoT over this baseline.
We further validated our framework in hardware experiments, and could successfully demonstrate robustness against disturbances, such as slipping. Significantly, we presented a motion planner that is easy to implement, computationally efficient, and saves energy by having the robot only make a step when necessary. Coupling of leg dynamics, other geometric shape representations, and providing further energy-efficient comparisons in hardware are left as future work.





 


\bibliography{bib.bib}

\begin{thebibliography}{10}

\bibitem{ranger_robot}
P.~A. Bhounsule, J.~Cortell, and A.~Ruina, ``Design and control of ranger: An energy-efficient, dynamic walking robot,'' {\em Adaptive Mobile Robotics}, pp.~441--448, 2012.

\bibitem{anymal_robot}
M.~Hutter {\em et~al.}, ``Anymal - a highly mobile and dynamic quadrupedal robot,'' in {\em 2016 IEEE IROS}, pp.~38--44, 2016.

\bibitem{schperberg2022auto}
A.~Schperberg, S.~D. Cairano, and M.~Menner, ``Auto-tuning of controller and online trajectory planner for legged robots,'' {\em IEEE Robotics and Automation Letters}, vol.~7, no.~3, pp.~7802--7809, 2022.

\bibitem{RL_energy}
Z.~Sun and N.~Roos, ``An energy efficient dynamic gait for a {Nao} robot,'' in {\em 2014 IEEE ICARSC}, pp.~267--272, 2014.

\bibitem{vision_energy}
L.~Chen, S.~Ye, C.~Sun, A.~Zhang, G.~Deng, T.~Liao, and J.~Sun, ``Cnns based foothold selection for energy-efficient quadruped locomotion over rough terrains,'' in {\em 2019 IEEE International Conference on Robotics and Biomimetics (ROBIO)}, pp.~1115--1120, 2019.

\bibitem{RoboImitationPeng20}
X.~B. Peng, E.~Coumans, T.~Zhang, T.-W.~E. Lee, J.~Tan, and S.~Levine, ``Learning agile robotic locomotion skills by imitating animals,'' in {\em Robotics: Science and Systems}, 07 2020.

\bibitem{Lee_2020}
J.~Lee, J.~Hwangbo, L.~Wellhausen, V.~Koltun, and M.~Hutter, ``Learning quadrupedal locomotion over challenging terrain,'' {\em Science Robotics}, vol.~5, no.~47, 2020.

\bibitem{Raibert}
M.~H. Raibert, {\em Legged Robots That Balance}.
\newblock USA: Massachusetts Institute of Technology, 1986.

\bibitem{kajita_ZMP}
S.~Kajita, F.~Kanehiro, K.~Kaneko, K.~Fujiwara, K.~Harada, K.~Yokoi, and H.~Hirukawa, ``Biped walking pattern generation by using preview control of zero-moment point,'' in {\em 2003 IEEE International Conference on Robotics and Automation}, vol.~2, pp.~1620--1626, 2003.

\bibitem{trajOpt}
A.~W. Winkler, C.~D. Bellicoso, M.~Hutter, and J.~Buchli, ``Gait and trajectory optimization for legged systems through phase-based end-effector parameterization,'' {\em IEEE Robotics and Automation Letters}, vol.~3, no.~3, pp.~1560--1567, 2018.

\bibitem{RL_2}
K.~Ito and F.~Matsuno, ``A study of reinforcement learning for the robot with many degrees of freedom - acquisition of locomotion patterns for multi-legged robot,'' in {\em 2002 IEEE International Conference on Robotics and Automation}, vol.~4, pp.~3392--3397, 2002.

\bibitem{imitation_deepMimic}
X.~B. Peng, P.~Abbeel, S.~Levine, and M.~van~de Panne, ``Deepmimic: example-guided deep reinforcement learning of physics-based character skills,'' {\em ACM Trans. Graph.}, vol.~37, July 2018.

\bibitem{energy_1}
H.-K. Shin and B.~K. Kim, ``Energy-efficient gait planning and control for biped robots utilizing the allowable {ZMP} region,'' {\em IEEE Transactions on Robotics}, vol.~30, no.~4, pp.~986--993, 2014.

\bibitem{energy_2}
L.~Chen, S.~Ye, C.~Sun, A.~Zhang, G.~Deng, T.~Liao, and J.~Sun, ``{CNNs} based foothold selection for energy-efficient quadruped locomotion over rough terrains,'' in {\em 2019 IEEE International Conference on Robotics and Biomimetics (ROBIO)}, pp.~1115--1120, 2019.

\bibitem{energy_4}
W.~Xi and C.~D. Remy, ``Optimal gaits and motions for legged robots,'' in {\em 2014 IEEE/RSJ International Conference on Intelligent Robots and Systems}, pp.~3259--3265, 2014.

\bibitem{Koco2014LocomotionCF}
A.~Mutka, E.~Ko{\v{c}}o, and Z.~Kova{\v{c}}i{\'c}, ``Adaptive control of quadruped locomotion through variable compliance of revolute spiral feet,'' {\em IJARS}, vol.~11, no.~10, p.~162, 2014.

\bibitem{model_based_energy1}
N.~Chakraborty, S.~Berard, S.~Akella, and J.~Trinkle, ``A geometrically implicit time-stepping method for multibody systems with intermittent contact,'' {\em IJRR}, vol.~33, no.~3, pp.~426--445, 2014.

\bibitem{model_based_energy2}
Z.~Manchester, N.~Doshi, R.~J. Wood, and S.~Kuindersma, ``Contact-implicit trajectory optimization using variational integrators,'' {\em IJRR}, vol.~38, no.~12-13, pp.~1463--1476, 2019.

\bibitem{fu2021minimizing}
Z.~Fu, A.~Kumar, J.~Malik, and D.~Pathak, ``Minimizing energy consumption leads to the emergence of gaits in legged robots,'' in {\em Conference on Robot Learning ({CoRL})}, 2021.

\bibitem{forceMPC}
J.~Di~Carlo, P.~M. Wensing, B.~Katz, G.~Bledt, and S.~Kim, ``Dynamic locomotion in the {MIT Cheetah 3} through convex model-predictive control,'' in {\em 2018 IEEE/RSJ IROS}, pp.~1--9, 2018.

\bibitem{schulman2017proximalpolicyoptimizationalgorithms}
J.~Schulman {\em et~al.}, ``Proximal policy optimization algorithms,'' {\em https://arxiv.org/abs/1707.06347}, 2017.

\bibitem{state_estimator}
M.~Menner and K.~Berntorp, ``Simultaneous state estimation and contact detection for legged robots by multiple-model kalman filtering,'' in {\em 2024 (ECC)}, pp.~2768--2773, 2024.

\bibitem{MELCHIORRI1994235}
C.~Melchiorri, ``Force manipulability ellipsoids for general manipulation systems,'' {\em IFAC Proceedings Volumes}, vol.~27, no.~14, pp.~235--240, 1994.
\newblock Fourth IFAC Symposium on Robot Control, Capri, Italy, September 19-21, 1994.

\bibitem{todorov2012mujoco}
E.~Todorov, T.~Erez, and Y.~Tassa, ``Mujoco: A physics engine for model-based control,'' in {\em 2012 IEEE/RSJ IROS}, pp.~5026--5033, IEEE, 2012.

\end{thebibliography}
\bibliographystyle{ieeetr}

\end{document}